\def\year{2021}\relax
\documentclass[letterpaper]{article} 
\usepackage{aaai21}  
\usepackage{times}  
\usepackage{helvet} 
\usepackage{courier}  
\usepackage[hyphens]{url}  
\usepackage{graphicx} 
\usepackage{natbib}  
\usepackage{caption} 
\usepackage{mathtools}
\usepackage{amsmath,amsthm,amssymb}
\usepackage{caption}
\DeclareCaptionFormat{myformat}{\fontsize{10.0}{10.0}\selectfont#1#2#3}
\title{Using Trust for Heterogeneous Human-Robot Team Task Allocation}

\author{
    Arsha Ali, Hebert Azevedo-Sa, Dawn M. Tilbury, Lionel P. Robert Jr.
    \\
}
\affiliations{
    University of Michigan \\
    Ann Arbor, MI 48109 \\
    \{arshaali, azevedo, tilbury, lprobert,\}@umich.edu

}

\begin{document}

\maketitle

\begin{abstract}
Human-robot teams have the ability to perform better across various tasks than human-only and robot-only teams. However, such improvements cannot be realized without proper task allocation. Trust is an important factor in teaming relationships, and can be used in the task allocation strategy. Despite the importance, most existing task allocation strategies do not incorporate trust. This paper reviews select studies on trust and task allocation. We also summarize and discuss how a bi-directional trust model can be used for a task allocation strategy. The bi-directional trust model represents task requirements and agents by their capabilities, and can be used to predict trust for both existing and new tasks. Our task allocation approach uses predicted trust in the agent and expected total reward for task assignment. Finally, we present some directions for future work, including the incorporation of trust from the human and human capacity for task allocation, and a negotiation phase for resolving task disagreements.
\end{abstract}

\section{Introduction}

Heterogeneous teams with human and robotic agents have the potential for better outcomes compared to homogeneous teams. Human-robot teams can work together towards a common goal by combining the strengths of both types of agents  \cite{hu_fatigue_allocation}. For example, the flexibility and reasoning of humans can be coupled with the speed and precision of robots \cite{hu_fatigue_allocation}.

In human teaming, trust is vital to ensuring proper task allocation in order to achieve optimal performance. When agents have an understanding of how much they can trust each other to execute a specific task, tasks can be appropriately allocated. When tasks are allocated optimally, overall team performance can be improved. We assume trust will also be important to task allocation in human-robot teaming, thereby reaching optimal levels of performance. 

This paper proposes a trust-based task allocation strategy for use in heterogeneous human-robot teams. To optimally allocate tasks, this strategy is based on trust in an agent and the expected total reward associated with the task and use of that agent. The contribution of this paper is a new task allocation strategy based on trust.

\section{Background and Related Studies}

Trust is an important concept in teaming relationships. Among trust definitions used in the literature, Lee and See define trust as ``the attitude that an agent will help achieve an individual’s goals in a situation characterized by uncertainty and vulnerability'' \cite{lee_see}. 
In a driver-automated vehicle (AV) setting, many factors influence driver trust in the AV including situational awareness \cite{mavric_augmented}, internal and external risk \cite{mavric_internal_external_risks}, and automation reliability \cite{mavric_false_alarms_misses}. 
Trust is also influenced by individual factors, such as age and gender \cite{ioe_trust_individual}. Researchers have used such findings to develop models to predict human trust in automation. \cite{hebertIJSR} used a Kalman filter to estimate driver trust in an AV based on driver behavior and performance on a non-driving task. \cite{optimo} created an online probabilistic trust inference model (OPTIMo), using a dynamic Bayesian network to estimate human trust in a robot based on robot performance and human intervention. \cite{soh} explored trust transfer across tasks through Bayesian and neural approaches where word-vector features were derived from English-language task descriptions.  

By predicting human trust in automation, automation can help calibrate trust to avoid the problematic cases of overtrust and undertrust. Overtrust can lead to misuse, where the trustor (the agent doing the trusting) relies on the trustee (the agent being trusted) to execute tasks beyond the trustee's capabilities, and undertrust can lead to disuse, where the trustor does not fully leverage the capabilities the trustee offers \cite{mavric_calibrate_trust}. One such study reduced trust miscalibrations by comparing driver's trust to AV capabilities, with the AV verbally communicating with the driver to modify trust \cite{mavric_calibrate_trust}.

The existing literature on task allocation has considered several approaches. \cite{alburaiki_specialized} introduced a probabilistic approach for assigning an agent from a non-homogeneous robotic swarm to a task based on the most qualified and available specialized agent that met a minimum fitting threshold as set by a human supervisor. However, \cite{alburaiki_specialized}'s approach does not consider an agent's performance history and is not applicable for new tasks, as tasks and agent specialties must be known before allocation. \cite{frame_allocation} developed an autonomous manager to dynamically redistribute tasks in a surveillance environment considering performance and human workload. Team performance and human workload are compared to user-specified thresholds to guide the logic for reallocation \cite{frame_allocation}. \cite{hu_fatigue_allocation} allocate tasks in a human-manufacturing system with the addition of modeling human fatigue in finite levels. They allocate tasks while minimizing an average human and operation cost \cite{hu_fatigue_allocation}. While \cite{frame_allocation} has the advantage of redistributing tasks in real-time and \cite{hu_fatigue_allocation} considers cost and changes in human fatigue levels, these approaches are difficult to be applied to new tasks without an estimation of agent performance on the new task. \cite{frame_allocation}'s approach relies on each agent's performance distribution for each task. For \cite{hu_fatigue_allocation}'s methodology, the probability that the agent will succeed on the task is required before allocation. 

Despite the importance of trust in human-robot teaming, existing task allocation techniques do not use trust between agents when assigning tasks. Research has shown that human trust transfers across tasks by similarity and difficulty \cite{soh}. Thus, if tasks are allocated by trust, unobserved tasks can be allocated. As demands are changing, robots are becoming more generalized with the ability to be repurposed and to handle various tasks, so the inclusion of trust has the potential to allocate new tasks.

\section{Bi-directional Trust Model (BTM)}
The BTM predicts both human trust in another agent, which we term \textit{natural trust}, and robotic trust in another agent, which we term \textit{artificial trust}. The BTM represents tasks as required capabilities and agents by their proven capabilities. Representing tasks and agents in this way allows for multi-task trust prediction to unobserved tasks, which are new tasks that the agent has not executed before. Also, when an agent's capabilities and task requirements are known, the trustor can adjust their trust in the trustee to execute the task by determining whether the trustee agent's capabilities are sufficient for the task requirements or not. 

Trustee capabilities are not known to the trustor in advance and are initialized from uniform belief distributions. Capability estimates are then built with observations as the trustee executes tasks, either as successes or failures. An agent's capabilities are therefore based on the history of their task outcomes. Trust is represented as the probability that a given agent can successfully execute a task, and is calculated by considering the task requirements and the agent's capabilities belief distribution. The model assumes that agents are more likely to be successful on tasks that have requirements less than the agent's capabilities. Similarly, agents are less likely to be successful on tasks that have capability requirements greater than the agent's capabilities. The trustee capabilities belief distribution, and hence the trust prediction, get better defined as more task outcomes are observed. 

The Cartesian product $\Lambda = \prod_{i=1}^n \Lambda_i = [0, 1]^n$ is used to represent $n$ distinct capabilities. A task is represented by $\gamma \in \Gamma$. An agent's capabilities are given by $\lambda = (\lambda_1, \lambda_2, ..., \lambda_n) \in \Lambda$ and the task capability requirements are given by $\bar{\lambda} = (\bar{\lambda}_1, \bar{\lambda}_2, ..., \bar{\lambda}_n) \in \Lambda$. 

\begin{figure}[t]
    \centering
    \includegraphics[width=1.0\columnwidth]{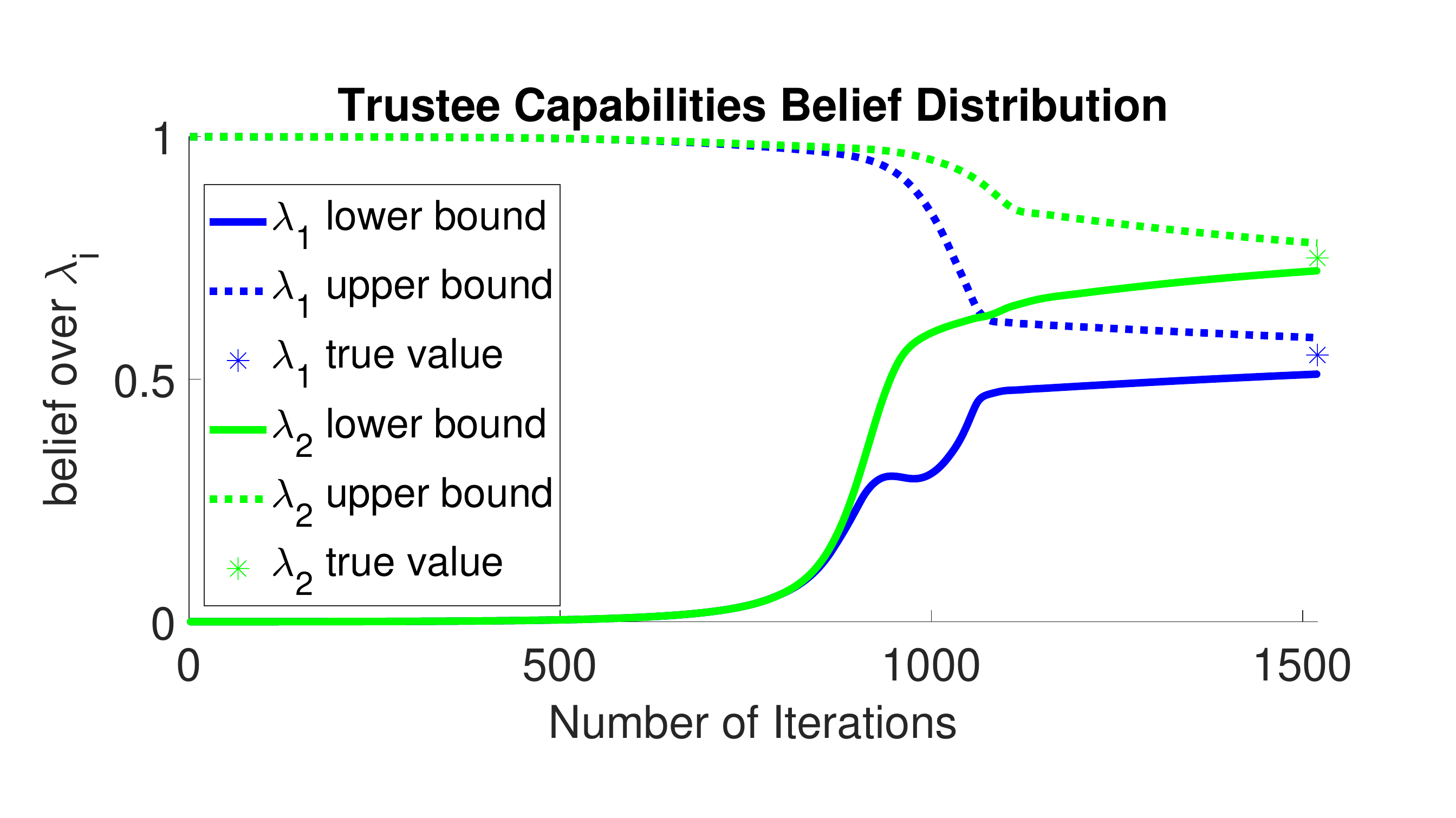}
    \caption{The update in belief over the trustee's $\lambda_1$ and $\lambda_2$ capabilities after observing $N = 800$ tasks executed by the trustee. The lower (solid lines) and upper bounds (dotted lines) for $\lambda_1$ (blue lines) and $\lambda_2$ (green lines) converge to the true $\lambda_1$ (blue asterisk) and $\lambda_2$ (green asterisk) values as the number of iterations increases.}
    \label{capabilityupdate}
\end{figure}

As an example in Figure \ref{capabilityupdate}, we updated the lower and upper bounds for a capability $\lambda_i$, $i \in \{1,2\}$ after observing N = 800 tasks executed by the trustee. For simplicity, we let the number of distinct capabilities $n = 2$ and $\lambda = (\lambda_1, \lambda_2) \in \Lambda$, where $\lambda_1$ and $\lambda_2$ are independent, unspecified capabilities. Each unspecified task $\gamma$ has some capability requirement $\bar{\lambda} = (\bar{\lambda}_1, \bar{\lambda}_2) \in \Lambda$. Initially, the capabilities of the trustee are unknown by the robot as indicated by lower bounds of $0$ and upper bounds of $1$ for a uniform distribution. After the robot observes the trustee's execution of these N = 800 tasks, the robot uses the artificial trust procedure to recursively update its belief in the trustee capabilities over 1520 iterations to generate the trustee capabilities belief that is most closely aligned with the observed outcomes. As the number of iterations increases, the capability belief distributions over $\lambda_1$ and $\lambda_2$ converge to the agent's true $\lambda_1$ and $\lambda_2$ capability values. The update in trustee capabilities could be done recursively after each task execution as well for real-world applications. More details about the BTM are available in \cite{mavric_BTM}.

\section{Trust-Based Task Allocation}

\begin{figure*}[t]
    \centering
    \includegraphics[width=1.0\textwidth]{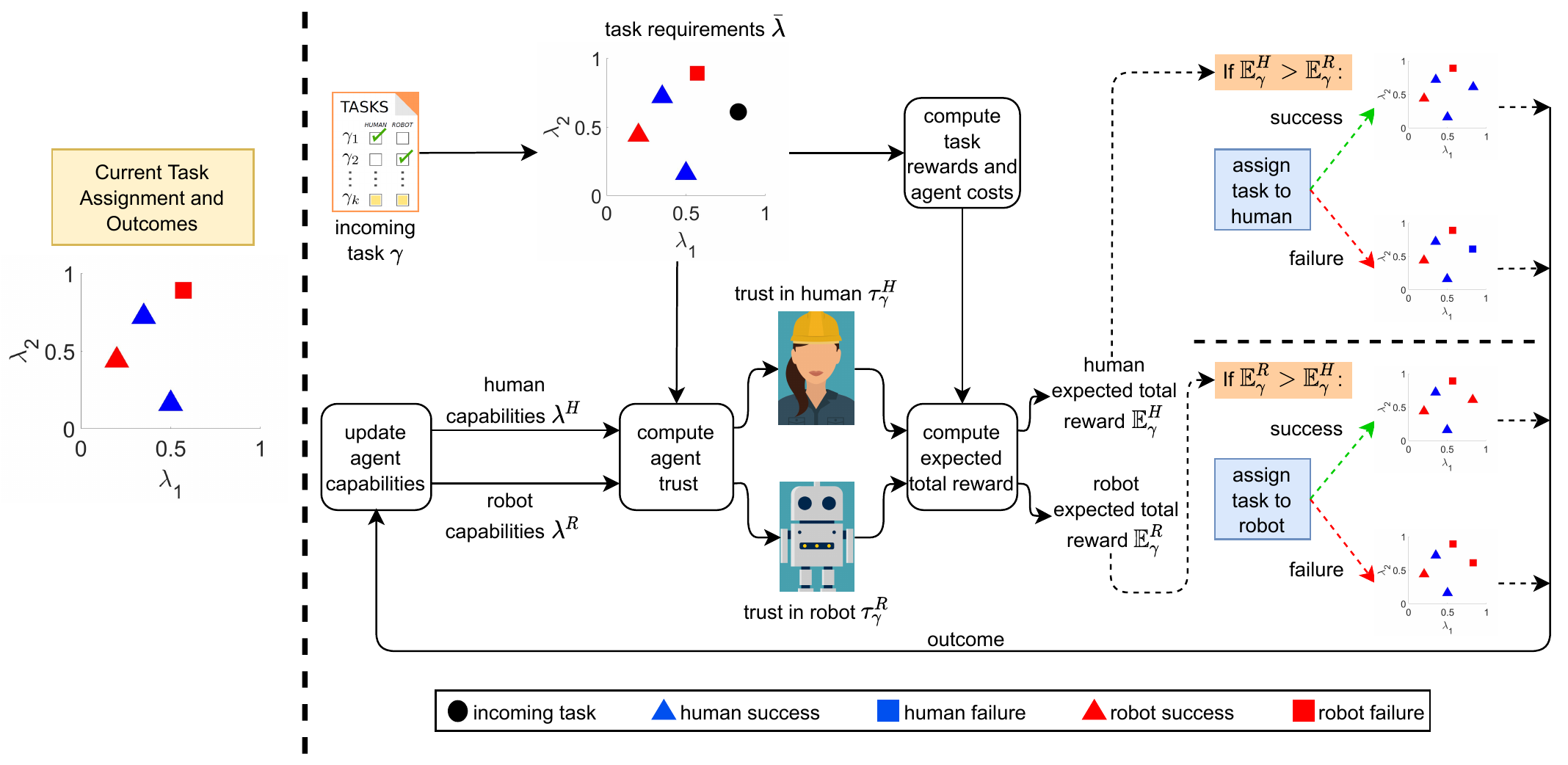} 
    \caption{The task allocation procedure for a team involving one human and one robotic agent. Task assignments (red for robot, blue for human) and outcomes (triangle for success, square for failure) are shown in the capability space $\Lambda$. An incoming task (black circle) with a set of capability requirements is used to compute task rewards, agent costs, and agent trust. An expected total reward is computed for each agent, and the task is assigned to the agent that maximizes the expected total reward. The outcome of that agent's task execution is observed and used to update the belief distribution for that agent's capabilities.}
    \label{blockdiagram}
\end{figure*}

The BTM can be used in a task allocation strategy for a heterogeneous human-robot team. The goal of the task allocation strategy is to assign each indivisible task (e.g., sorting one item from a conveyor belt) to an agent on the team. We propose to do this through the agent's expected total reward by balancing trust in the agent, reward for task success, penalty for task failure, and cost associated with using the agent for the given task. The task will be assigned to the agent that maximizes the expected total reward function from the robot's perspective, with the aim of improving overall team performance. The challenge is to define the functions, where the expected total reward function will depend on trust in the agent and the functions for reward for success, penalty for failure, and agent cost. 

Figure \ref{blockdiagram} shows a flowchart of the task allocation procedure for a team with one human and one robotic agent. The process begins with an incoming task $\gamma$ that needs to be executed. The task is mapped to a set of capability requirements $\bar{\lambda}$, which we assume are embedded with the task. The task requirements are then used to calculate the task reward and agent costs, as well as trust in each agent. The reward for success and penalty for failure both depend on the task requirements. The agent cost depends on the task requirements and on the potential trustee agent who may execute the task. Trust in each agent from the robot's perspective is computed using the task requirements and capabilities of the agent. Each agent's expected total reward is then computed, and the agent that maximizes the expected total reward is assigned the task. The outcome of the task execution is observed as either a success or a failure and used to update the trustee agent's capabilities belief distribution.

A possible expected total reward equation $\mathbb{E}_\gamma^a[total\;reward]$ for an agent $a \in \{{H, R}\}$ ($H$ represents the human agent and $R$ represents the robotic agent) is given by Eq. (\ref{totalReward_eqn}), \begin{equation}\label{totalReward_eqn}
\mathbb{E}_\gamma^a[total\;reward] = {\tau}_\gamma^a(r_{s}-c^a) + (1-{\tau}_\gamma^a)(r_{f}-c^a) ,
\end{equation}
where trust ${\tau}_\gamma^a$ is the probability of agent $a$ successfully executing the task $\gamma$ and $1-{\tau}_\gamma^a$ is the probability of failing as outputted by the BTM. The reward for success $r_s$ captures the importance of executing the task successfully, while the penalty for failure $r_f$ captures the punishment for failing at the task. The cost $c^a$ is the cost of using agent $a$ for the task.

Assuming $r_f = 0$, Eq. (\ref{totalReward_eqn}) can be reduced to Eq. (\ref{totalReward_eqn2}),
\begin{equation}\label{totalReward_eqn2}
\mathbb{E}_\gamma^a[total\;reward] = {\tau}_\gamma^ar_{s}-c^a .
\end{equation}

One way to define the function for reward for success $r_s$ is to depend on the task requirements $\bar{\lambda}$, i.e., $r_s = f_r(\bar{\lambda})$, $\bar{\lambda} \in [0, 1]^n$. The reward for success $r_s$ will be such that tasks with higher requirements will yield a greater value than tasks with lower requirements. A potential definition of cost for an agent $c^a$ to execute the task can depend on the type of agent, either a human or a robot, and the task requirements, i.e., $c^a = f_c(a,\bar{\lambda})$, $a \in \{{H, R}\}$. The agent cost $c^a$ can be greater for tasks with higher requirements, assuming that such tasks require greater resources from the agent. Also, the agent cost $c^a$ for a given task can always be greater for a human agent than a robotic agent, capturing that a human agent is more flexible and can take on a larger variety of tasks than a robot.

There are many options for structuring and defining an expected total reward equation from which to allocate tasks. A simple manifestation of the functions for reward for success $r_s$, human agent cost $c^H$, and robotic agent cost $c^R$ can be through a weighted linear combination of the task requirements $\bar{\lambda}$. It is important to note that under the logic presented, a more interesting case of task assignment occurs when the human and robot excel in distinct capability dimensions. If the capabilities of the robot exceed the capabilities of the human, the task will probably be assigned to the robot since the cost of the robot is likely to be less than the cost of the human. In principle, it is possible that the expected total reward between agents is the same, although it should be rare. In such a case, the task can be assigned to either agent, perhaps considering which agent has fewer tasks already assigned to it.

\section{Future Work}

We presented a method for task allocation based on trust. In the future, we plan to include trust from the human's perspective and determine how long it takes to learn trustee capabilities. Also, one idea to expand this framework is to consider human capacity while allocating tasks. Capacity is envisioned to be a general concept, capturing elements that affect task performance such as emotion and workload. In a takeover transition study, a calm emotion resulted in a higher takeover readiness than anger \cite{ioe_emotions_short}. Performance can also be boosted by circumventing the workload cases of underload and overload \cite{widyanti_gsrconcept}. Underload can cause boredom and decrease performance, and overload can lead to fatigue, increase the risk of errors, and result in injury and accident \cite{widyanti_blinkrate, widyanti_flightattendant}. The problem will involve estimating human capacity in real-time and characterizing how changes in capacity should affect the allocation of tasks. 

\begin{figure}[h]
    \centering
    \includegraphics[scale = 0.73]{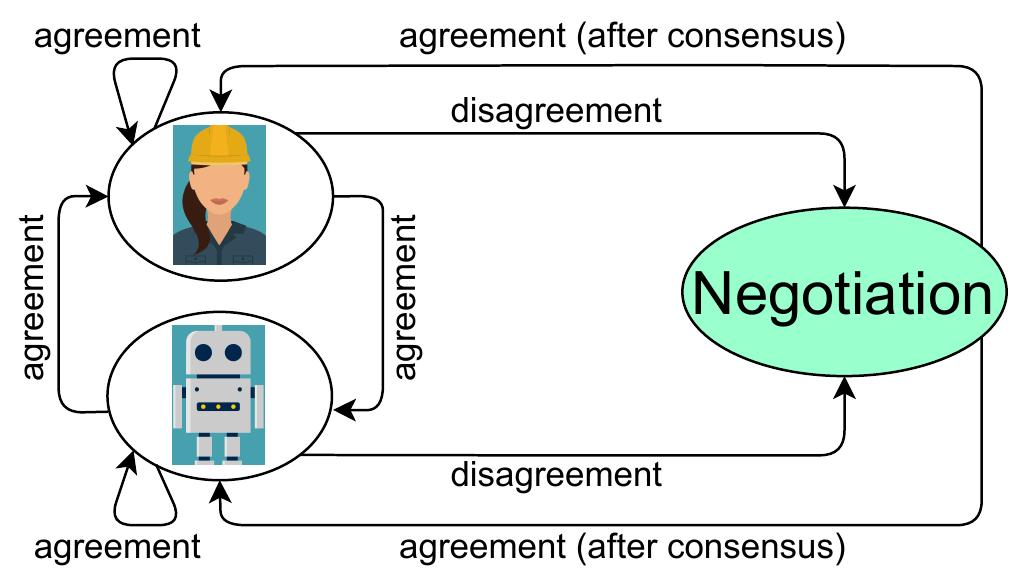}
    \caption{Task allocation map including a negotiation phase between a human and a robot.}
    \label{negotiationphase}
\end{figure}

Another direction for future work is to investigate how humans and robots can negotiate and reach a decision on the assignment of a task if there is disagreement between the agents on who should do the task. Figure \ref{negotiationphase} shows the process of assigning tasks between a human and a robot including the negotiation phase. If the agents are in agreement on which agent should execute the task for best performance, there is no need for negotiation. However, when agents have differing opinions, one agent may interrupt another agent as it executes the task it was assigned, potentially causing frustration and confusion in human agents, reducing the collaborative team relationship, and decreasing team performance. Thus, the ability for agents to negotiate fairly and reach a decision on task assignment when there is disagreement is important. When there is disagreement, the agents can negotiate until reaching a consensus.

\section{Acknowledgements}
This work is funded in part by the National Science Foundation and the Army Research Lab under grant \#F061352.

\fontsize{9.0pt}{10.0pt} \selectfont
\bibliography{ref.bib}
\end{document}